\documentclass[sigconf,natbib=false]{acmart}
\usepackage[english]{babel}
\usepackage[backend=bibtex]{biblatex}
\usepackage{colortbl}
\usepackage{csquotes}
\usepackage{cleveref}

\AtBeginDocument{%
  \providecommand\BibTeX{{%
    \normalfont B\kern-0.5em{\scshape i\kern-0.25em b}\kern-0.8em\TeX}}}

\makeatletter
\patchcmd{\maketitle}
  {\andify\authors}
  {\add@english@keywords\andify\authors}
  {}{}
\def\englishkeywords#1{%
	\if\relax\detokenize{#1}\relax
		\gdef\add@english@keywords{}%
	\else
		\gdef\add@english@keywords{%
			\@specialsection{Keywords}#1\par
		}%
    \fi
}
\makeatother

\setcopyright{rightsretained}
\copyrightyear{2022}
\acmDOI{}

\acmPrice{}
\acmISBN{}
\begin{document}

\title[Unified Question Answering in Slovene]{Unified Question Answering in Slovene}

\author{Katja Logar and Marko Robnik-Šikonja}
\affiliation{
\institution{University of Ljubljana, Faculty of Computer and Information Science}
\city{Ljubljana}
\country{Slovenia}
}
\email{kl2164@student.uni-lj.si,marko.robnik@fri.uni-lj.si} 




\begin{abstract}
  Question answering is one of the most challenging tasks in language understanding. Most approaches are developed for English, while less-resourced languages are much less researched. 
  We adapt a successful English question-answering approach, called UnifiedQA, to the less-resourced Slovene language. Our adaptation uses the encoder-decoder transformer SloT5 and mT5 models to handle four question-answering formats: yes/no, multiple-choice, abstractive, and extractive. We use existing Slovene adaptations of four datasets, and machine translate the MCTest dataset. We show that a general model can answer questions in different formats at least as well as specialized models. The results are further improved using cross-lingual transfer from English. While we produce state-of-the-art results for Slovene, the performance still lags behind English.

  
\end{abstract}

\englishkeywords{question answering, Slovene language, deep neural networks, encoder-decoder models, natural language processing}


\maketitle

\section{Introduction}

Most studies for the question answering (QA) task deal with the English language. This leaves many language specifics, not present in English, potentially inadequately addressed. E.g., some problematic language specifics in morphologically-rich Slovene language are noun and adverb declension, three different genders, three counts, the person or pronoun being hidden in a verb, etc. An additional problem for less-resourced languages is the lack of suitable datasets for QA.

Khashabi et al. \cite{khashabi2020unifiedqa} argue that building specialized models for each QA dataset or QA format is unnecessary, as they all require a similar inference capability. Therefore, it is possible to develop one model capable of answering questions in different formats. They call their approach UnifiedQA, and we adapted this approach to Slovene. 

The number of QA datasets in Slovene is much lower than used in the original UnifiedQA. We found four partially human-translated but mostly machine-translated datasets. To improve that, we first machine translate the additional MCTest dataset \cite{richardson-etal-2013-mctest} into Slovene and fix translation errors. 

Our method is based on the pretrained Slovene encoder-decoder transformer model SloT5 \cite{ulcar2022sequence}. We finetune the model on the five QA datasets and analyze its performance.
We also test the role of uppercase and lowercase letters, the impact of unanswerable questions, and the contribution of each dataset to the performance of the unified model. Next, we test the cross-lingual transfer and train a multilingual question answering model based on the multilingual mT5 model \cite{xue2021mt5}, using English and Slovene datasets. Finally, we perform a qualitative analysis of the obtained models.
The results show that our system is currently the best performing QA system for Slovene. We make its source code freely accessible\footnote{ \href{https://github.com/klogar/QAslovene}{https://github.com/klogar/QAslovene}}.

The paper is split into four further sections. In \Cref{sec:related}, we outline the related work on QA in Slovene. \Cref{sec:methodology} presents our adaptation of UnifiedQA methodology and the applied Slovene QA datasets, and \Cref{sec:experiments} discusses different evaluation settings and their results. In \Cref{sec:conclusions}, we present the findings and ideas for further improvements.

\section{Related work}
\label{sec:related}
The QA in Slovene is relatively unexplored. In the pre-neural setting, Čeh et al. \cite{vceh2009slovene} developed a closed-domain QA system for answering common questions that arise during students' studies at the University of Maribor, Faculty of Electrical Engineering, Computer Science and Informatics. The translation of the SuperGLUE benchmark suite 
to Slovene in 2021 \cite{vzagar2022slovene} provided four partially human, partially machine translated QA datasets (BoolQ, COPA, MultiRC, and ReCoRD) and evaluation of Slovene BERT models. 
Ulčar et al. \cite{ulcar2022sequence} adapted the SloT5 model for the yes-no and multiple-choice questions. Finally, Zupanič et al. \cite{zupanic2021cross} translated the SQuAD 2.0 dataset from English and adapted different multilingual models. They achieved the best result with the SloBERTa 2.0 model \cite{ulvcar2021sloberta}. In contrast to the above works, we apply the transfer learning paradigm within the encoder-decoder SloT5 and mT5 models and provide a unified approach to different QA formats, obtaining the best results so far.


\section{Methodology}
\label{sec:methodology}
Our methodology follows Khashabi et al. \cite{khashabi2020unifiedqa} UnifiedQA methodology.
The authors define four QA formats (extractive, abstractive, multiple-choice, and yes/no) and unify the learning approach to these formats. The extractive format requires that the answer is directly stated in the supplied context as a substring. The abstractive format requires paraphrasing of the given context and the answer may require linking information from several sentences. The multiple-choice datasets have possible answers listed and the aim is to select the given option correctly. Finally, the yes/no questions require only yes or no as an answer. 

The datasets with different QA formats are converted to text format, with parts of the input separated by the \verb#"\n"# separator. Extractive, abstractive and yes/no questions are coded as  \verb#"question \n context"#  and multiple-choice questions as \\ \verb#"question \n possible choices \n context"#. Here, the possible choices are indicated in capital letters from A onwards \verb#(A) choice 1# \verb#(B) choice 2...#. 


We initially considered four QA datasets. Three stem from the translation of the SuperGLUE benchmark to Slovene \cite{vzagar2022slovene}: MultiRC \cite{khashabi2018looking} (abstractive), COPA \cite{roemmele2011choice} (multiple-choice) and BoolQ \cite{clark-etal-2019-boolq} (yes/no). We also used the SQuAD 2.0 \cite{rajpurkar2018know} (extractive) Slovene translation \cite{zupanic2021cross}. SQuAD 2.0 contains unanswerable questions, and some are also present in MultiRC. As we focus on the reading comprehension task, all selected datasets have a context. COPA is a commonsense reasoning dataset, which is not our primary focus, but we included it due to being human translated into Slovene. BoolQ, MultiRC, and SQuAD 2.0 are partially human translated \cite{vzagar2022slovene, zupanic2021cross}. 

To have a non-commonsense multiple-choice dataset, we machine translated the MCTest dataset \cite{richardson-etal-2013-mctest} and fixed some translation errors. To reduce the cost of translation, we partially used the commercial solution DeepL \cite{deepl2022} and partially an internal neural machine translator of a bit lesser quality. Later, we translated the entire MCTest dataset with the DeepL translator and made it publicly available in our repository. However, the reported results are obtained using the initial mixed translation setting.

As the starting training model for monolingual Slovene UnifiedQA models, we used the monolingual Slovene variant of the T5 transformer encoder-decoder model \cite{raffel2020exploring}, called SloT5 \cite{ulcar2022sequence}. For the cross-lingual transfer experiments, we applied the multilingual variant of T5, called mT5 \cite{xue2021mt5}. Due to computational time and GPU memory limitations, we used the SloT5 and mT5 models of the smallest size (60M and 300M parameters, respectively). Originally, \citeauthor{khashabi2020unifiedqa} used the T5 model \cite{raffel2020exploring} of the largest possible size (11B parameters) and the BART\textsubscript{large}  model \cite{lewis2020bart} as a starting point for the UnifiedQA model. However, they also report results for the T5\textsubscript{small} model, which we report for comparison, so all models are of comparable sizes. Table \ref{tab:params} lists the parameters used to finetune our models.

\begin{table}[htb]
  \caption{Parameters for finetuning UnifiedQA models. }
  \label{tab:params}
  \begin{tabular}{lc}
    \toprule
    Parameter & Value \\
    \midrule
    Maximum input size [tokens] & 512 \\
    Maximum output size [tokens] & 100 \\
    Number of epochs & 25 \\
    Batch size & 8 \\ 
    Number of beams & 4 \\
    Learning rate & 5e-5 \\
  \bottomrule
\end{tabular}
\end{table}

\section{Experiments and results}
\label{sec:experiments}
In this section, we report our work on empirical evaluation. We present the evaluation metrics, original English results, experiments and results in the monolingual Slovene setting, and in the cross-lingual transfer setting.

\subsection{Evaluation Metrics}
\label{sec:metrics}
For each dataset, we use a different evaluation metric. For BoolQ, we report the classification accuracy; for SQuAD 2.0, the $F_1$ score; for MultiRC, we use ROUGE-L; and for the multiple-choice datasets (MCTest and COPA), we calculate the best match between the generated text and the offered options and compute the classification accuracy. In all cases, the answers are first normalized (removing punctuation and unnecessary spaces and converting the text to lowercase). 



\subsection{English UnifiedQA Results Using T5\textsubscript{small}}
First, we replicated the results of the original English UnifiedQA \cite{khashabi2020unifiedqa} and also obtained the results for the datasets not originally used, i.e. COPA and MultiRC (the latter was only used as a yes/no dataset in \cite{khashabi2020unifiedqa}). The results are presented in Table \ref{tab:engmodel}. The results for BoolQ and MCTest are slightly worse than originally reported, which could be attributed to slightly different parameters for text generation. We achieved a much worse result for the SQuAD 2.0 dataset, with $F_1$ only 46.1\% rather than 67.6\%. Trying to replicate the published scores with the original code\footnote{\href{https://github.com/allenai/unifiedqa) }{https://github.com/allenai/unifiedqa}}, we obtained similar results to ours . However, we analyzed the difference and believe that at least some of them are due to unanswerable questions, as the $F_1$ score is 84.5\% for questions that have an answer and only 7.8\% for unanswerable questions. The UnifiedQA model, therefore, does a poor job of detecting if a question is unanswerable from the context.

\begin{table}[htb]
  \caption{Our and published results of the UnifiedQA (UniQA) approach on English datasets using the T5\textsubscript{small} model. }
  \label{tab:engmodel}
  \resizebox{\columnwidth}{!}{
  \begin{tabular}{ l | c c c c c c c c c }
    \toprule
Dataset     & BoolQ & COPA & MCTest & MultiRC & SQuAD 2.0 \\
Metric    & CA & CA & CA & ROUGE-L & $F_1$ & \\
    \midrule
    UniQA(publ.) & 0.771 & / & 0.800 & / & 0.676 \\
    UniQA(ours) & 0.757 & 0.560 & 0.762 & 0.536 & 0.461 \\  \bottomrule
\end{tabular}
}
\end{table}

\subsection{Slovene Monolingual Results Using SloT5}
In the Slovene monolingual setting, we compare different variants of Slovene UnifiedQA models and report the results in Table \ref{tab:generalmodel}.
We adapted the models for each QA format separately and obtained so-called specialized models. These provided a baseline for what could be achieved with each individual QA format. We then trained the SloUnifiedQA model using all available Slovene datasets. We also investigated the impact of unanswerable questions (SloUnifiedQA-NA, SloUnifiedQA-NA2, explained below) and the use of only lower case letters (SloUnifiedQA-LC). 

\begin{table}[htb]
  \caption{Comparing variants of Slovene UnifiedQA approach (based on the SloT5 model). Besides the unified model, we report the results of specialized models for each QA format (specialized),  the best results published so far on these datasets  (published), and the default classifier. The effect of unanswerable questions and lowercasing is analyzed in the bottom part of the table. Note that SloUniQA-NA is tested on modified datasets without unanswerable questions, so the results for this model are incomparable. }
  \label{tab:generalmodel}
  \resizebox{\columnwidth}{!}{
  \begin{tabular}{ l | c c c c c c c | c}
    \toprule
Dataset     & BoolQ & COPA & MCTest & MultiRC & SQuAD 2.0 & \\
Metric    & CA & CA & CA & ROUGE-L & $F_1$ & Avg.\\
    \midrule
        SloUniQA & 0.683 & 0.532 & 0.463 & 0.310 & 0.555 & 0.509 \\
        specialized & 0.688 & 0.486 & 0.439 & 0.255 & 0.554 & 0.484 \\
        published & 0.666 & 0.500 & / & / & 0.739 & / \\
        default  & 0.623 & 0.500 & 0.269 & / & / & / \\
        \hline
        
        SloUniQA-NA & 0.675 & 0.524 & 0.454 & 0.319 & 0.637 & 0.522 \\
        SloUniQA-NA2 & \textbf{0.695} & \textbf{0.554} & \textbf{0.474} & \textbf{0.321} & \textbf{0.556} & \textbf{0.520} \\
        SloUniQA-LC & 0.686 & 0.530 & 0.449 & 0.259 & 0.533 & 0.491 \\
  \bottomrule
\end{tabular}
}
\end{table}


Comparing the SloUnifiedQA model with specialized models, the models achieve better results for the multiple-choice datasets (COPA and MCTest) and the abstractive dataset (MultiRC). The improvement for the extractive dataset is minimal, and we observe a slight decrease in accuracy for the yes/no dataset (BoolQ). Better results are also obtained compared to all main classifiers.

Comparing SloUnifiedQA on Slovene with the English UnifiedQA model on English datasets (in \Cref{tab:engmodel}), the English model gives better results for all selected formats except SQuAD 2.0. Interestingly, the English and Slovene models have different problems with SQuAD 2.0. The Slovenian one predicts unanswerable questions too often (it has $F_1$ score of 60,3\% for unanswerable questions and only 50,4\% for answerable ones, while incorrectly identifying 13\% of answerable questions as unanswerable), the English one too rarely. At the same time, the English model never wrongly predicts that a question is unanswerable. This is likely due to unanswerable questions making up a larger proportion of the dataset in the Slovene training dataset than in the English one. For other datasets, the biggest difference in metrics can be observed in the MCTest multiple-choice dataset, where the difference is 33\%. We attribute the worse result of SloUnifiedQA to machine translations and a much smaller training dataset, especially for the multiple-choice questions; as in the original work, the authors use three additional datasets in addition to MCTest.

Compared to other published works on the same datasets, we achieve better results with the SloUnifiedQA on the BoolQ and COPA datasets compared to \citeauthor{ulcar2022sequence} \cite{ulcar2022sequence}, while on the SQuAD 2.0 dataset, Zupanič et al. \cite{zupanic2021cross} achieve a significantly better result (almost 20\%). Here, \citeauthor{ulcar2022sequence}  \cite{ulcar2022sequence} also use the SloT5 model with the textual output, while Zupanič et al. \cite{zupanic2021cross} use the SloBERTa model and only predict the span of the answer, which is an easier task.


\subsubsection{The Effect of Unanswerable Questions} \ \\
Unanswerable questions account for about one-third of all training examples, and models could overfit such questions. To address this issue, we train two models, SloUnifiedQA-NA and SloUnifiedQA-NA2. For the SloUnifiedQA-NA model, we removed all unanswerable questions. As evident from \Cref{tab:generalmodel}, for yes/no questions and multiple-choice questions the accuracy deteriorates, while for abstractive and extractive questions the metrics improve. The biggest improvement occurred for the SQuAD 2.0 dataset, where the $F_1$ metric for answerable questions improved to 63.7\%. 


The SloUnifiedQA-NA was the basis for the SloUnifiedQA-NA2 model, which we trained on complete datasets, including unanswerable questions. The metrics slightly improved for BoolQ, COPA, and MCTest but may be due to the longer training time. No improvement is observed for SQuAD 2.0; the $F_1$ for answerable questions even drops to 51.5\%.

\subsubsection{The Effect of Using Lower Case Letters} \ \\
To analyze the effect of using only lower case letters, we trained the SloUnifiedQA-LC model. The results are comparable for BoolQ and COPA, but for MCTest, MultiRC, and SQuAD 2.0, the results are worse. The uppercase letters, therefore, contain relevant information in Slovene.

\subsubsection{Contribution of Datasets in the Unified Model} \ \\
To assess the impact of each dataset in the SloUnifiedQA model, we dropped each training dataset in turn. The results are shown in Table \ref{tab:ablation}. The largest individual performance drop is observed for the model without BoolQ, as the yes/no questions become unanswerable (the CA for the BoolQ dataset is almost 0\%). This also strongly affects the average impact but causes even slight improvements on MCTest, MultiRC, and SQuAD 2.0. The second largest average performance drop is achieved by the model without SQuAD 2.0, where a drop is observed on all datasets. For other models, the drops are observed mainly on datasets on which models were not trained. Overall, the COPA dataset contributes the least to the performance of SloUnifiedQA, the corresponding model achieving almost the same performance.

\begin{table}[htb]
\caption{Contribution of datasets in the unified model by omitting one dataset at a time. The red color indicates the two largest performance drops for each dataset.}
\label{tab:ablation}
\resizebox{\columnwidth}{!}{
\begin{tabular}{ l | c c c c c c c | c}
 \toprule
Dataset  & BoolQ & COPA & MCTest & MultiRC & SQuAD2.0 & \\
Metric & CA & CA & CA & ROUGE-L & $F_1$ & Avg.\\
 \midrule
SloUniQA & 0.683 & 0.532 & 0.463 & 0.310 & 0.555 & 0.509 \\
\hline
no BoolQ & \cellcolor{red!45} 0.001 & 0.522 & 0.486 & 0.319 & 0.561 & 0.378 \\
no SQuAD 2.0 & \cellcolor{red!20} 0.664 & \cellcolor{red!20} 0.516 & \cellcolor{red!20} 0.451 & \cellcolor{red!20} 0.258 & \cellcolor{red!45} 0.120 & 0.402 \\
no MCTest & 0.676 & \cellcolor{red!45} 0.510 & \cellcolor{red!45} 0.351 & 0.317 & 0.560 & 0.483 \\
no MultiRC & 0.690 & 0.536 & 0.457 & \cellcolor{red!45} 0.209 & \cellcolor{red!20} 0.552 & 0.489 \\
no COPA & 0.683 & \cellcolor{red!45} 0.510 & 0.456 & 0.319 & 0.554 & 0.504 \\
 \bottomrule
\end{tabular}
}
\end{table}


\subsection{Cross-Lingual Transfer Using mT5}
There are only a few QA datasets in Slovene, so we checked if using transfer from additional English datasets can improve the Slovene results. We used three different collections of datasets. 
\begin{itemize}
    \item \textbf{SLO}: Slovene datasets BoolQ, COPA, MCTest, MultiRC and SQuAD 2.0 (described in \Cref{sec:methodology}).
    \item \textbf{ANG5}: English datasets BoolQ, COPA, MCTest, MultiRC, and SQuAD 2.0 (the English dataset, whose translations form the SLO collection).
    \item \textbf{ANG9}: English datasets BoolQ, COPA, MCTest, MultiRC, and SQuAD 2.0 and all datasets, used by Khashabi et al. \cite{khashabi2020unifiedqa}, except SQuAD 1.1, i.e. NarrativeQA, RACE, ARC, and OBQA. 
\end{itemize}

We trained five models using the multilingual mT5 model on these dataset collections and tested them on the SLO test sets.
The first model, mSloUnifiedQA, was trained only on SLO datasets and gives a baseline performance of mT5, also enabling comparison to monolingual SloT5. The mSloUnifiedQA\textsubscript{1} models were trained on both English and Slovene datasets simultaneously (only one phase), with the English dataset collection being either ANG5 or ANG9. Only the SLO dataset group was used for validation. The mSloUnifiedQA\textsubscript{2} models were trained in two phases, first on the English datasets (ANG5 or ANG9), using the ROUGE-L metric to select the best model, and the obtained model was then finetuned on the SLO dataset collection.

\begin{table}
\caption{Results of cross-lingual transfer using additional English datasets and multilingual models based on mT5.}
\label{tab:munified}
\resizebox{\columnwidth}{!}{
\begin{tabular}{ l | c c c c c c c | c}
 \toprule
Dataset & BoolQ & COPA & MCTest & MultiRC & SQuAD 2.0 &  \\
Meric & CA & CA & CA & ROUGE-L & $F_1$ & Avg. \\
\midrule
mSloUniQA & 0.646 & 0.488 & 0.515 & 0.298 & 0.571 & 0.504 \\
mSloUniQA\textsubscript{1} (ANG5) & 0.672 & 0.486 & 0.582 & 0.308 & 0.587 & 0.527 \\
mSloUniQA\textsubscript{1} (ANG9) & 0.676 & 0.508 & 0.579 & \textbf{0.340} & 0.598 & \textbf{0.540} \\
mSloUniQA\textsubscript{2} (ANG5) & 0.682 & 0.504 & 0.564 & 0.313 & 0.593 & 0.531 \\
mSloUniQA\textsubscript{2} (ANG9) & \textbf{0.683} & 0.486 & \textbf{0.602} & 0.323 & \textbf{0.604} & \textbf{0.540} \\
 \bottomrule
\end{tabular}
}
\end{table}

The results of the five multilingual models are presented in Table \ref{tab:munified}. Comparison between the monolingual SloUnifiedQA model (in \Cref{tab:generalmodel}) and the multilingual mSloUnifiedQA shows that they perform on average equally well, with SloUnifiedQA performing better on the BoolQ, COPA and MultiRC datasets, and mSloUnifiedQA performing better on the MCTest and SQuAD 2.0 datasets. 

Adding additional knowledge in English improved the average metrics by 3-4\%, but the training time increased by about four times for the models with the most datasets (ANG9). A slight improvement can be observed for models using nine English datasets (ANG9) relative to those with only five English datasets (ANG5). The additional datasets contribute the most to the MCTest multiple-choice results, but the performance on MultiRC and SQuAD 2.0 also improved. On the other hand, despite the additional datasets, the results for BoolQ and COPA are worse than for the monolingual model. Using one or two-phase training does not make a difference on average, but there are differences in individual datasets.

\subsection{Qualitative Analysis}
Qualitative analysis of our models showed that the generated answers are mostly substrings or given choices in multiple-choice questions. Models cannot paraphrase, rephrase or provide answers in the correct Slovene case. They also have problems with multi-part questions requiring multiple answers that are not listed in the same place in the context. Machine translations, which are not always grammatically correct or do not make it clear what the question is asking for, also make answering the questions difficult. The models performed best on factoid questions that require a short answer.

\section{Conclusion and future work}
\label{sec:conclusions}

The main contributions of this work are the generative unified QA models based on SloT5 and mT5 encoder-decoder transformer models, which set new state-of-the-art results for QA in Slovene. An additional contribution is the machine-translated and corrected MCTest dataset. 

We identify three possible directions for further work. First, better translations or dedicated Slovenian datasets would improve upon currently mainly machine-translated datasets. Second, larger T5 models and longer training times have shown better performance in English. In our work, we used only the smallest available T5 models due to the limited memory of the GPU; we also limited training sessions to a maximum of 25 epochs. Third, by using new datasets, especially additional multiple-choice datasets, as evidenced by the improvement brought by the introduction of English multiple-choice datasets. Further, additional abstractive datasets could teach the models to rephrase better or that answers shall not be just substrings of the provided context.

\section*{Acknowledgments}
Marko Robnik-\v Sikonja received financial support from the Slovenian Research Agency through core research programme P6-0411 and projects J6-2581 and J7-3159, as well as the Ministry of Culture of Republic of Slovenia through the project Development of Slovene in Digital Environment (RSDO).

\printbibliography

\appendix









\end{document}